\documentclass[]{ceurart}

\usepackage{listings}
\usepackage{hyperref}
\usepackage{placeins} 

\begin{document}

\copyrightyear{2026}
\copyrightclause{Copyright for this paper by its authors.
  Use permitted under Creative Commons License Attribution 4.0
  International (CC BY 4.0).}
\conference{Preprint}

\title{Concept-based explanations of Segmentation and Detection models in Natural Disaster Management}

\author[1]{Samar Heydari}
\address[1]{Department of Artificial Intelligence, Fraunhofer Heinrich Hertz Institute, Berlin, Germany}

\author[1]{Jawher Said}

\author[1]{Galip Ümit Yolcu}

\author[1]{Evgenii Kortukov}

\author[1]{Elena Golimblevskaia}

\author[2]{Evgenios Vlachos}[%
orcid=0009-0002-2247-2124
]
\address[2]{Department of Informatics, Aristotle University of Thessaloniki, Thessaloniki, Greece}

\author[2]{Vasileios Mygdalis}[%
orcid=0000-0001-5473-5262
]

\author[2]{Ioannis Pitas}[%
orcid=0009-0006-7555-8641
]

\author[1]{Sebastian Lapuschkin}[%
orcid=0000-0002-0762-7258,
email=sebastian.lapuschkin@hhi.fraunhofer.de
]
\cormark[1]

\author[1]{Leila Arras}[%
email=leila.arras@hhi.fraunhofer.de
]
\cormark[1]

\cortext[1]{Corresponding author.}

\begin{abstract}
Deep learning models for flood and wildfire segmentation and object detection enable precise, real-time disaster localization when deployed on embedded drone platforms. However, in natural disaster management, the lack of transparency in their decision-making process hinders human trust required for emergency response. To address this, we present an explainability framework for understanding flood segmentation and car detection predictions on the widely used PIDNet and YOLO architectures. More specifically, we introduce a novel redistribution strategy that extends Layer-wise Relevance Propagation (LRP) explanations for sigmoid-gated element-wise fusion layers. This extension allows LRP relevances to flow through the fusion modules of PIDNet, covering the entire computation graph back to the input image. Furthermore, we apply Prototypical Concept-based Explanations (PCX) to provide both local and global explanations at the concept level, revealing which learned features drive the segmentation and detection of specific disaster semantic classes. Experiments on a publicly available flood dataset show that our framework provides reliable and interpretable explanations while maintaining near real-time inference capabilities, rendering it suitable for deployment on resource-constrained platforms, such as Unmanned Aerial Vehicles (UAVs).
\end{abstract}


\maketitle

\section{Introduction}

Natural disasters such as flood and wildfire pose a serious threat to public safety and the global economy \cite{owid-natural-disasters}, requiring advanced monitoring systems. Deep neural networks (DNNs) for the semantic segmentation and the detection of objects from drone and satellite images can provide crucial information for situational awareness. 
However, their “black-box” nature limits the trust and confidence required for emergency responses. We address this challenge by introducing an end-to-end explainable framework for DNN-based segmentation and object detection in natural disaster management (NDM) building upon Layer-wise Relevance Propagation (LRP) \cite{bach2015lrp}, an explainable AI attribution technique that provides feature-level importance scores (aka relevances) to explain individual predictions of neural networks.
Together with its concept-based extensions, Concept Relevance Propagation (CRP) \cite{Achtibat2023,DreCVPRW23} and Prototypical Concept-based Explanations (PCX) \cite{dre2024c}, LRP provides an understanding of DNN model behavior both globally (i.e., dataset-wide) and in terms of human-comprehensible concepts. 

LRP and its concept-based extensions already demonstrated their usefulness in computer vision, but they have not yet been applied to a PIDNet architecture nor to the NDM domain. Additionally, Prototypical Concept-based Explanations (PCX) \cite{dre2024c} were so far confined to image \textit{classification} models: we present its first extension to semantic \textit{segmentation} and object \textit{detection} models.

\textbf{Contributions.} In summary, our novel contributions are as follows:
\begin{itemize}
    \item We apply LRP \cite{bach2015lrp} and concept-based explanations \cite{Achtibat2023,DreCVPRW23} for the first time in NDM. 
    \item We extend LRP \cite{bach2015lrp} to a PIDNet architecture.
    \item We apply PCX \cite{dre2024c} for the first time on segmentation and detection models.
\end{itemize}

\section{Related Work}

\subsection{Deep learning for Natural Disaster Management}

Deep learning methods are increasingly common in Natural Disaster Management (NDM) \cite{AKHYAR2024112067,make4020020} for real-time monitoring and response planning. In the context of wildfire, deep learning has been employed for satellite-based \cite{fire8020036} and aerial-based image analysis \cite{LI2022145}. Recent solutions leverage high-performance architectures like PIDNet \cite{PIDNet}, with specialized extensions for fusing infrared and RGB spectra such as RoboFireFuseNet \cite{Fotiou_2025}, and Neural Architecture Search (NAS) methods \cite{11226416} for optimizing the number of trainable parameters. Regarding flood segmentation, new learning processes utilize Self-Knowledge Distillation (Self-KD) to encode foreground information while suppressing background noise \cite{11084431}. Core enabling resources include publicly available datasets for wildfire/burned area segmentation such as Blaze \cite{11084700}, datasets and benchmarks for flood segmentation \cite{11044745}, as well as synthetic data \cite{fire_scientific_report, spatharis2025unrealfire} and weakly supervised learning frameworks \cite{tzimas2026extreme, APOSTOLIDIS2026133290} to address data scarcity and improve model generalization. Recent efforts regarding object detection in NDM employed transfer learning on YOLO models \cite{PI2020101009} and visibility-enhanced models, such as VE-DINO \cite{smartcities8010012}, to handle occlusion-prone environments. Other research focused on combining deep learning vision models with social media analysis \cite{wieland2025fusion} and sentiment analysis \cite{Alqithami2025} to give first responders a holistic understanding of the situation, which can help them allocate resources and better predict the progression of the disaster.

\subsection{Explainable AI}

Explainable AI (XAI) methods aim to achieve fidelity (explanations that align with the actual logic of the model) and comprehensibility for human operators. One popular class of XAI methods include local feature attributions that explain individual predictions by assigning importance scores to the input and latent features. These can be roughly divided into three groups: 1) perturbation/surrogate-based (e.g., SHAP\cite{Lundberg:NIPS2017}), 2) gradient-based (e.g., Gradient and Grad-CAM\cite{Simonyan:ICLR2014,Selvaraju:ICCV2017}), and 3) decomposition-based methods (e.g., LRP \cite{bach2015lrp}). Group 1 and 2 present the advantage of being model-agnostic. While group 1 are computationally expensive, group 2 are cheap  but typically noisy and prone to gradient shattering. Group 3 methods can be made as efficient as a gradient backward pass \cite{arras2025}, however they require a careful design for new types of neural network layers. Evaluation in controlled environments w.r.t. ground truths has demonstrated the superior quality of the latter group of methods \cite{arras2022clevr,Mamalakis2022}.

\subsection{Concept-based explanations}

Concept Relevance Propagation (CRP) \cite{Achtibat2023,DreCVPRW23} breaks down decisions into human-interpretable concepts by considering feature maps inside a convolutional neural network (CNN) as semantic concept detectors and conditioning the LRP backward pass on these concepts, making it possible to visualize concept-conditioned heatmaps in the input space, as well as retrieve samples that maximize the relevance of a concept. Prototypical Concept-based Explanations (PCX) \cite{dre2024c} generalize this to global XAI by summarizing the model's prediction behavior dataset-wide into prototypes. Another concept-based explanation is Testing with Concept Activation Vectors (TCAV) \cite{KimWGCWVS18} which measures the gradient-based sensitivity of the model's latent features w.r.t. pre-defined concept directions which are obtained by training a linear classifier to separate latent features of samples with and without that concept.

\section{Methods}

\subsection{Layer-wise Relevance Propagation (LRP)}

Layer-wise Relevance Propagation (LRP) \cite{bach2015lrp}  is a post-hoc, model-specific explanation technique that performs a conservative backward decomposition of the model's prediction, for instance, an object detection score or a semantic segmentation logit, backward through all the layers of the network until the input. The distribution of the relevance is controlled by local propagation rules that preserve the total relevance at each layer, i.e., $\sum_{i} R_{i}^{(l)} = \sum_{j} R_{j}^{(l+1)}$, where $R_{i}^{(l)}$ is the relevance of neuron $i$ in layer $l$. The basic LRP $\epsilon$-rule for a linear layer with neurons $i$ and $j$ in consecutive layers is given by:

\begin{equation}
\label{eps_rule}
R_i^{(l)} = \sum_j \frac{z_{ij}}{\sum_{k} z_{kj} + \epsilon \cdot \operatorname{sign}\left(\sum_{k} z_{kj}\right)} R_j^{(l+1)}
\end{equation}
In this equation, $z_{ij}$ is the contribution of neuron $i$ to neuron $j$ in the forward pass (typically equal to the neuron's activation $x_i$ multiplied by the connection weight $w_{ij}$, i.e., $z_{ij} = x_i \cdot w_{ij}$), and $\epsilon$ is a small numerical stabilizer. This conservative redistribution guarantees that the total evidence for a detection or segmentation is taken into account throughout the whole computation graph. Other LRP rules for CNNs include the $z^{+}$-rule and $\gamma$-rule. For an overview of LRP rules we refer to \citet{Montavon:ExplAIBook2020}.

\subsection{Extending LRP to PIDNet}
The PIDNet architecture introduces special layers which require propagation rules consistent with the LRP conservative backward decomposition. In particular, residual summations and bilinear interpolations are treated as linear layers and explained with the $\epsilon$-rule from Eq. \eqref{eps_rule}.

Additionally,  for element-wise multiplications of branches, as they occur in the PIDNet Pixel-Attention-Guided
(Pag) and Boundary-Attention-Guided (Bag) fusion layers, of the form $y = x \odot \sigma(g)$, where $\sigma$ is the sigmoid activation function, we propose to follow the signal-take-all redistribution strategy introduced for gated interactions in LSTMs \cite{arras2017lstm}, i.e., relevance is assigned entirely to the signal branch while the gating branch does not receive any relevance:
\begin{equation}
R_{x} = R_{y}, \qquad R_{g} = 0,
\end{equation}
reflecting the interpretation that the gated input $\sigma(g)$ acts only as a modulator of the signal $x$ in the forward pass; its effect is therefore already reflected in the relevance $R_y$.

\subsection{Concept-based explanations for Segmentation and Detection}

To leverage LRP latent feature attributions into concept-based explanations for segmentation and detection, building upon Concept Relevance Propagation (CRP) \cite{Achtibat2023,DreCVPRW23} and Prototypical Concept-based Explanations (PCX) \cite{dre2024c}, we proceed in the following way.
For each prediction of a segmentation mask or of an object bounding box, we start by generating latent LRP relevances of feature maps inside convolutional layers. These relevances are summed up across spatial dimensions to obtain \textit{concept relevance vectors} with one value per feature map. Then, in a second step, we cluster these vectors for all training samples using Gaussian Mixture Model (GMM) clustering. Each resulting cluster then represents a model prediction \textit{strategy}. Then, at inference time, when a new test prediction is made, we compare it to the nearest cluster centroid, also called a \textit{prototype}, in terms of concept usage. This enables us to quantify how similar or dissimilar a new test prediction is w.r.t. prototypical decisions, and whether the prediction shall be labeled as ordinary or as an outlier by PCX.

Besides, we visualize the semantic of concepts  through the retrieval of concept maximizing reference samples over the training data, together with generating concept-conditioned heatmaps of the prediction.
Example PCX prototypes, concepts and heatmaps in NDM will be provided in Section~\ref{section:prototypes}~and~\ref{section:single_predictions}.

\begin{figure}[h]
\centering
\begin{minipage}{0.49\textwidth}
\centering
\includegraphics[width=\linewidth, trim=8mm 0cm 0cm 9mm, clip]{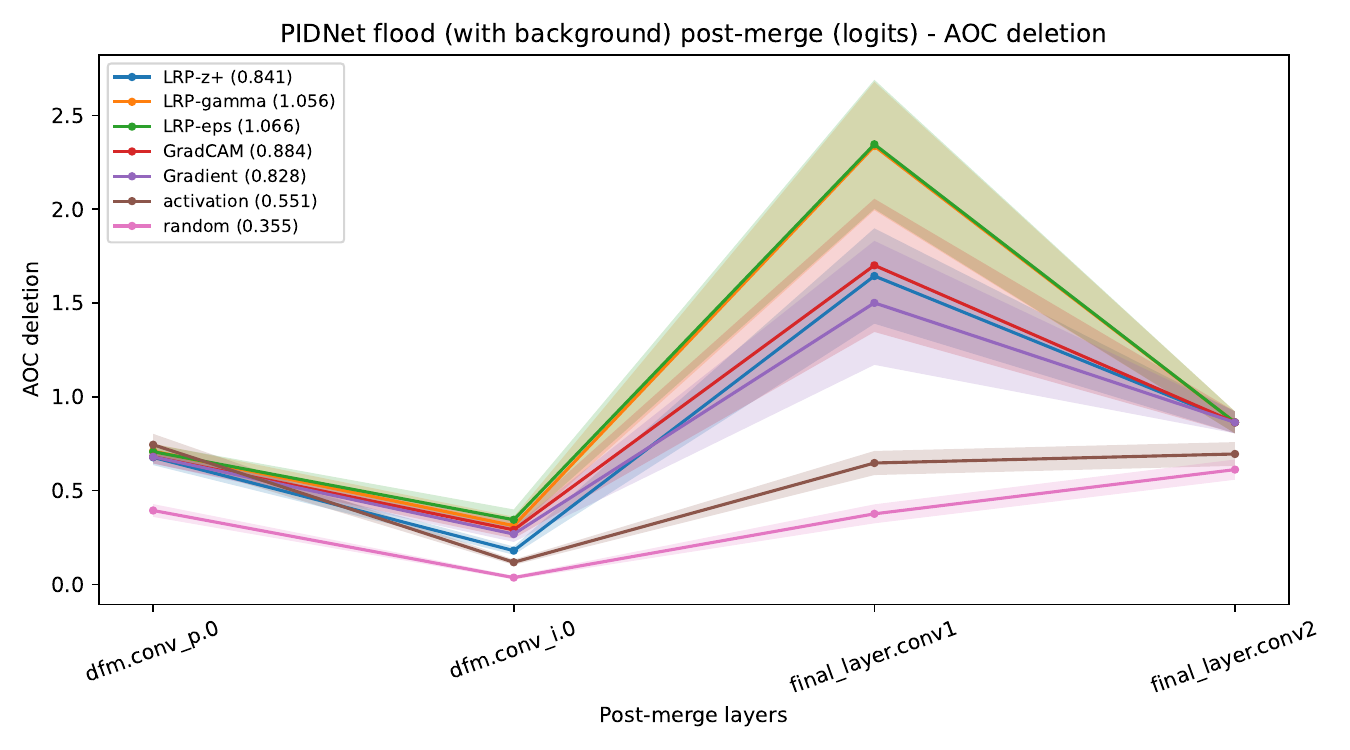}
\end{minipage}
\hfill
\begin{minipage}{0.49\textwidth}
\centering
\includegraphics[width=\linewidth, trim=8mm 0cm 0cm 9mm, clip]{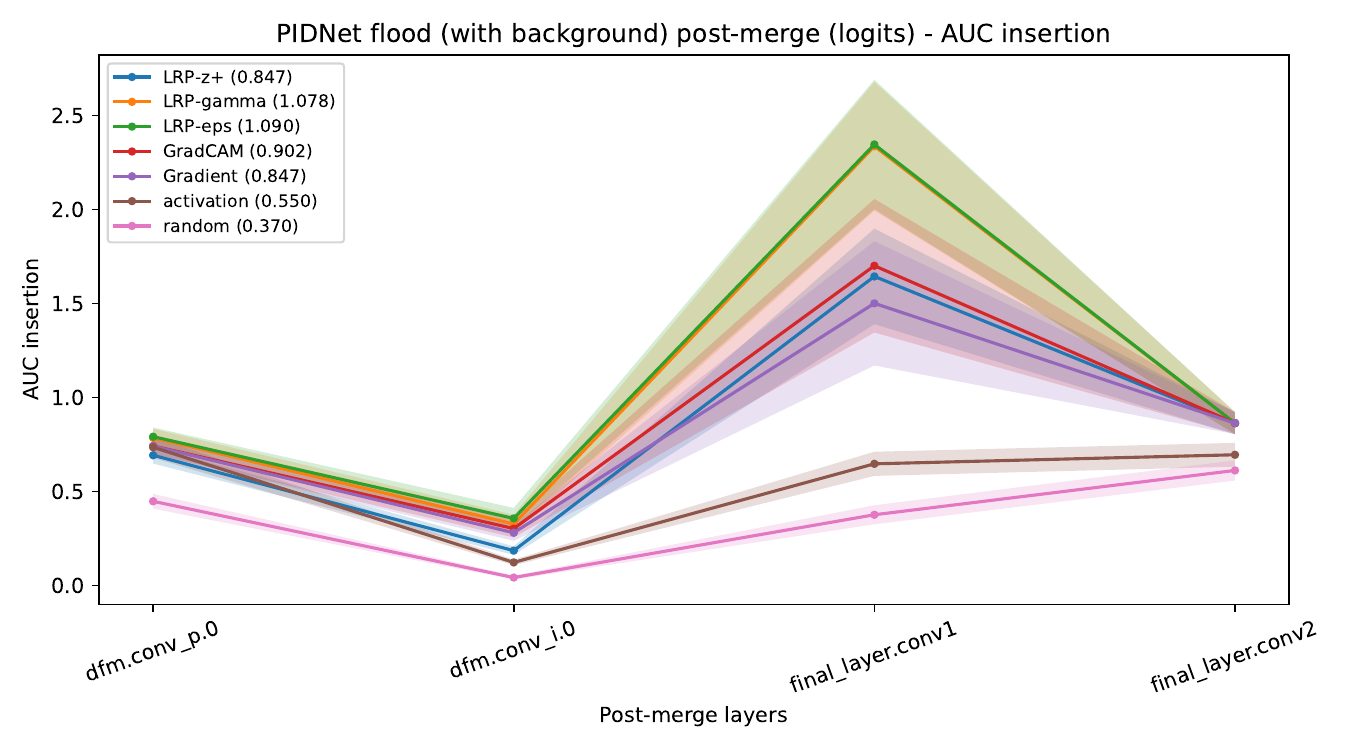}
\end{minipage}
\vspace{0.2cm}
\begin{minipage}{0.48\textwidth}
\centering
\includegraphics[width=\linewidth, trim=8mm 0cm 0mm 9mm, clip]{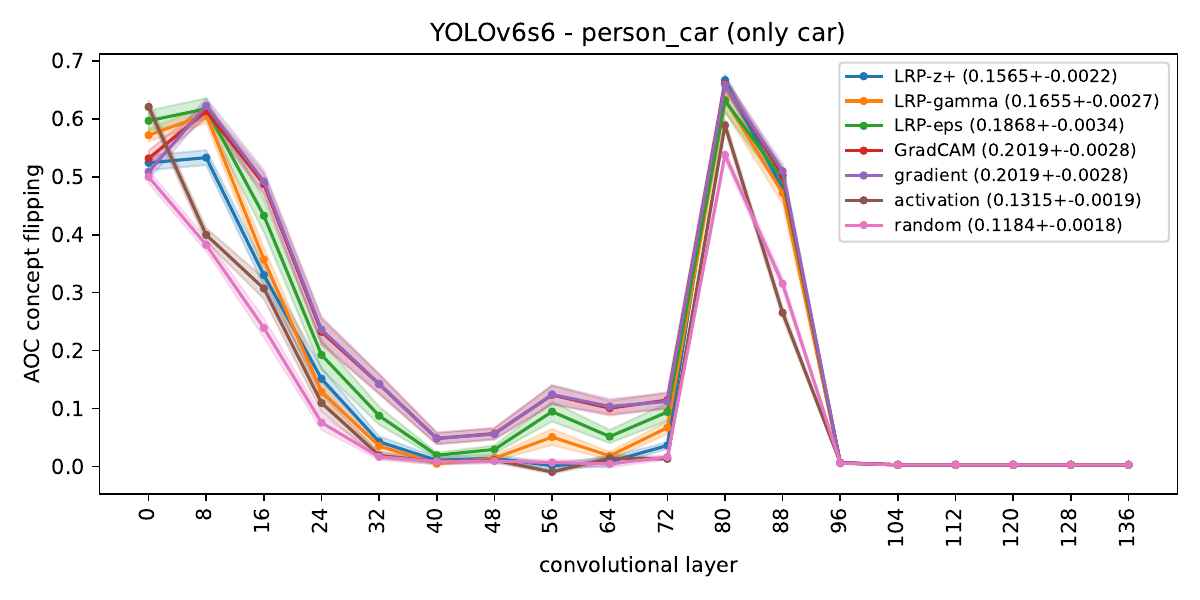}
\end{minipage}
\hfill
\begin{minipage}{0.48\textwidth}
\centering
\includegraphics[width=\linewidth, trim=8mm 0cm 0cm 9mm, clip]{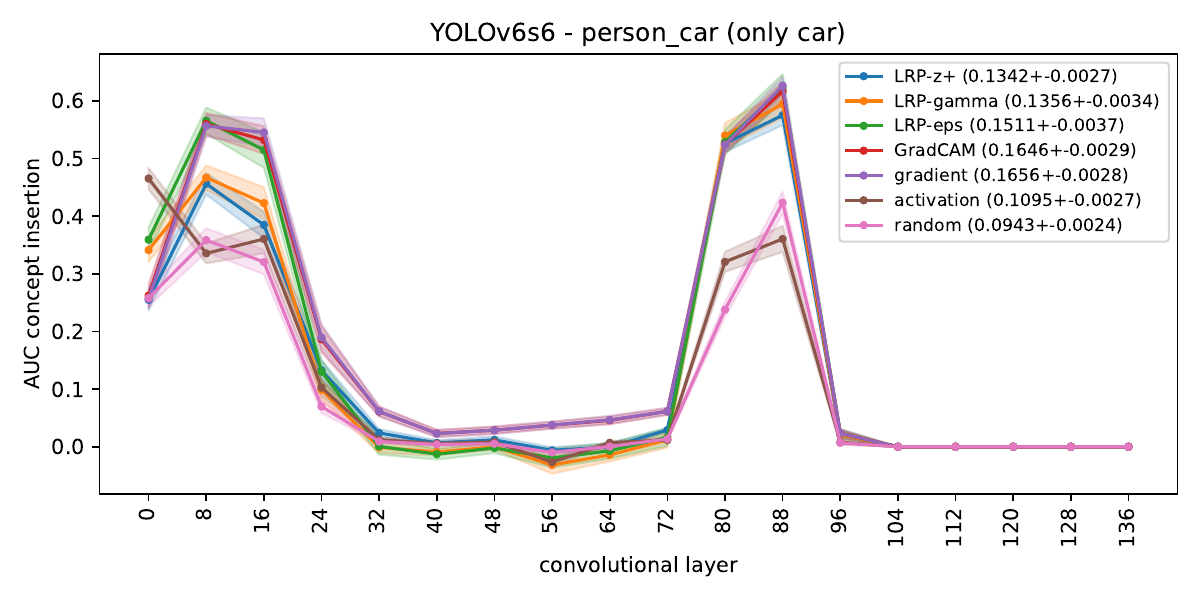}
\end{minipage}
\caption{Perturbation-based evaluation of concept-based explanations. \textit{Top}: PIDNet flood segmentation, \textit{Bottom}: YOLOv6s6 car detection. \textit{Left}: AOC for concept deletion, \textit{Right}: AUC for concept insertion. The higher the AOC/AUC scores, the better. AOC/AUC scores averaged over all layers are given in parenthesis.}\label{fig:perturbation}
\end{figure}

\section{Experimental results}\label{section:experiments}

\subsection{Setting: data and models}

We test our XAI framework on two state-of-the-art DNNs trained for flood segmentation and car/person detection in a flood scenario using the TEMA AIIA Ahrtal flood dataset\footnote{Dataset available at: \url{https://doi.org/10.5281/zenodo.18377521}} captured from UAVs.

\paragraph{PIDNet-small \cite{PIDNet}} This segmentation DNN is made of a  three-branch architecture inspired by control theory whose Pixel-Attention-Guided (Pag) and Boundary-Attention-Guided (Bag) fusion modules aggregate information from the detail (P-branch), context (I-branch), and boundary branches (D-branch). It consists of 27 layers and 8.06M parameters. The prediction performance over 244 validation images is: 0.833 mIoU, 0.91 Pixel Accuracy. The number of training images is 1321.

\paragraph{YOLOv6s6 \cite{Li2024YOLOv6}} This object detection DNN employs an EfficientRep backbone for feature
extraction and a Rep-PAN neck for multi-scale feature fusion, enabling the detection of objects across
various scales in a single stage. It consists of 136 layers and 41.32M parameters. The prediction performance over 179 validation images is: 0.428 F1@0.5, 0.255 mAP@0.5. Since the performance on the car class alone is higher: 0.571 F1@0.5, resp. 0.425 mAP@0.5, we focus on car detection in our experiments. The number of training images is 435 (pre-training was performed on the VisDrone Dataset).

\subsection{Evaluation of concept-based explanations}

In order to quantitatively assess concept-based explanations, we measure the Area Over the Curve (AOC), resp. Area Under the Curve (AUC), in the prediction logit \textit{change} when deleting, resp. inserting, feature maps according to their relevance over 100 samples from the validation data (the feature map relevance is obtained by summing up LRP relevances across spatial dimensions). This is analog to pixel-flipping for input feature relevance evaluation \cite{samek_2017}. For PIDNet we perform this perturbation-based evaluation only on the last 4 convolutional layers (i.e., after the 3 branches of PIDnet are merged, since it is unclear how to evaluate concepts in parallel layers), while for YOLOv6s6 we consider all 136 convolutional layers of the model (in steps of 8). Results are provided in Fig.~\ref{fig:perturbation}. We find that explanations based on LRP-$\epsilon$ \cite{bach2015lrp}, Gradient \cite{Simonyan:ICLR2014} and Grad-CAM \cite{Selvaraju:ICCV2017} are superior, while explanations using standard Activation are only slightly better than Random, which is consistent with previous work \cite{DreCVPRW23}. Among the LRP rules, we note that although LRP-$\epsilon$ performs better in the present perturbation-based evaluation, other evaluations taking into account object localization or ground truth masks have shown that rules such as $z^{+}$-rule should generally be preferred in computer vision CNNs \cite{arras2022clevr,Kohlbrenner_2020}.

\subsection{Visualizing prototypes and concepts}\label{section:prototypes}

\begin{figure}[h]
  \centering
  \includegraphics[width=0.6\linewidth, trim=0mm 0cm 0cm 4mm, clip]{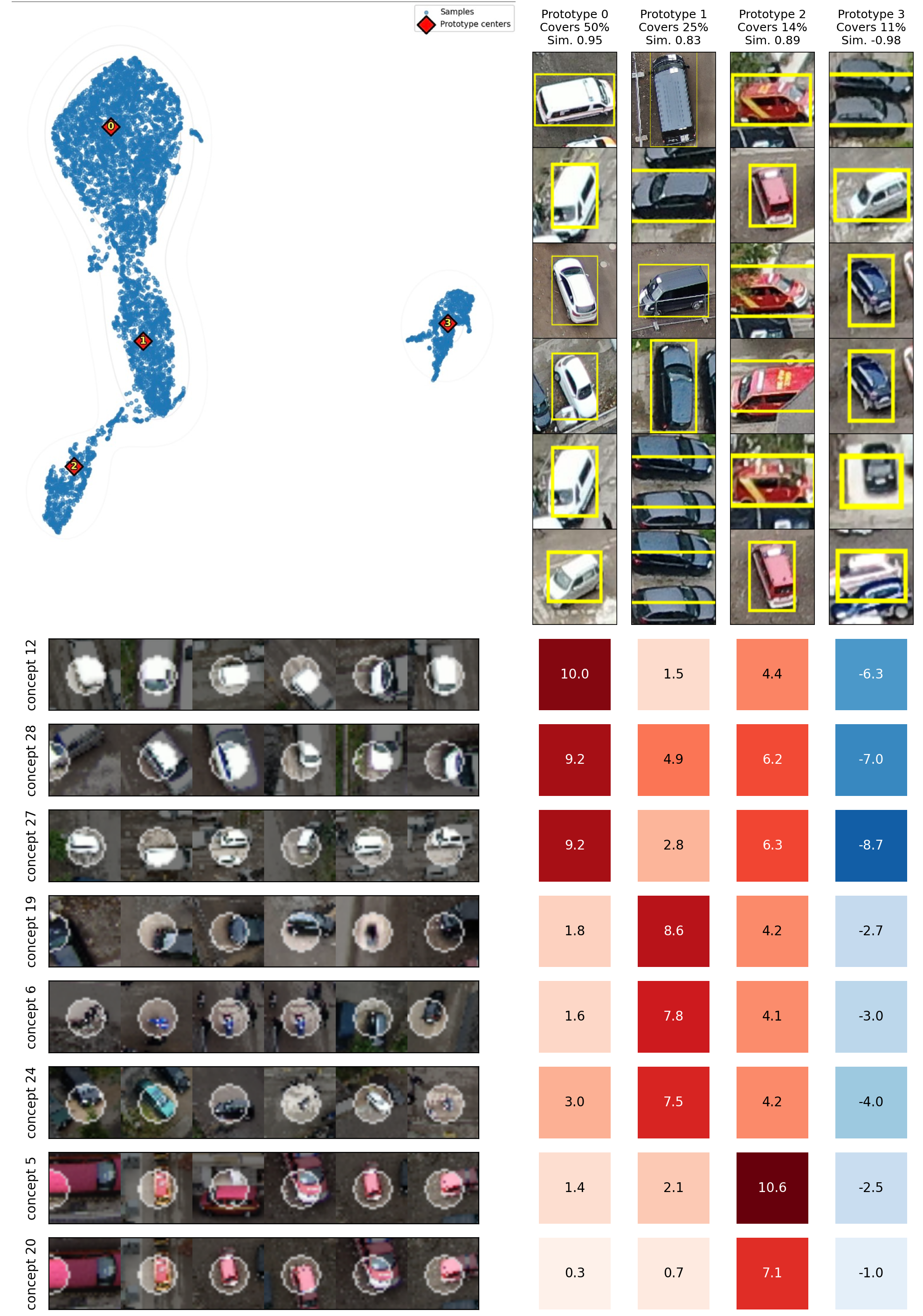}
  \caption{PCX prototypes, and their concept contributions, for car detection with YOLOv6s6.}\label{fig:grid_plot_YOLO}
\end{figure}

\begin{figure}[h]
  \centering
  \includegraphics[width=0.855\linewidth, trim=17mm 0cm 0cm 0mm, clip]{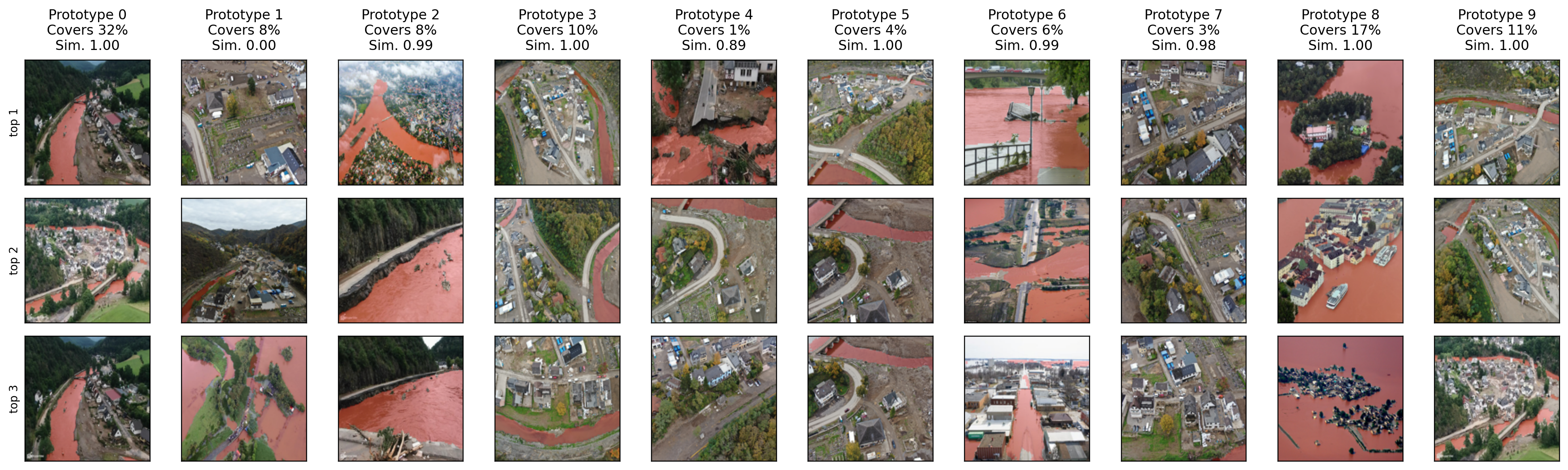}
  \caption{PCX prototypes for flood segmentation with PIDNet.}\label{fig:prototype_plot_PIDNet}
\end{figure}

We visualize PCX prototypes, and most relevant concepts, for car detection on the YOLOv6s6 training data in Fig.~\ref{fig:grid_plot_YOLO} (using layer \texttt{backbone.stem.rbr\_dense.conv} and 4 clusters for GMM clustering). We observe that the model has developed 4 different strategies to detect cars in a flood scenario, although it was trained to recognize all types of cars altogether. Prototypes 0 and 1 correspond to white and dark passenger cars resp., prototype 2 consists of red ambulances, while the remaining prototype 3 groups various blurry cars. Accordingly, predictions of each prototype rely on concepts of vehicle parts of the same color, except for prototype 3 where common concepts are under-used. In Fig.~\ref{fig:grid_plot_YOLO} we further indicate the coverage of each prototype in the data, and the cosine similarity of the \textit{concept relevance vector} of each cluster w.r.t. the full data. Prototype 3 covers 11\% of the training samples, and has a negative similarity to the car's mean of -0.98, reinforcing a data quality issue for this subset of samples. These PCX insights can serve to warn the end-user when a new test prediction is made which gets assigned to prototype 3, indicating that the prediction is less reliable and that the car type is atypical for the model.

In Fig.~\ref{fig:prototype_plot_PIDNet} we visualize the prototypes for flood segmentation on the PIDNet training data (using layer \texttt{final\_layer.conv1} and 10 clusters). We observe that the model has learned to distinguish different flood patterns. For example, prototypes 0, 3 and 9 correspond to linear flood structures, while prototype 8 represents wide-area plain inundation and prototype 6 captures small-scale, street-level flooding. Prototypes 1 and 4 with lower similarity to the flood's mean likely indicate outlier clusters with atypical flood patterns. Concepts in flood segmentation mainly correspond to water colors and floods occurring near vegetation, roads, or habitation (we refrain from retrieving such concepts due to space constraints).

\subsection{Understanding an individual prediction}\label{section:single_predictions}

\begin{figure}[h]
  \centering
  \includegraphics[width=0.84\linewidth]{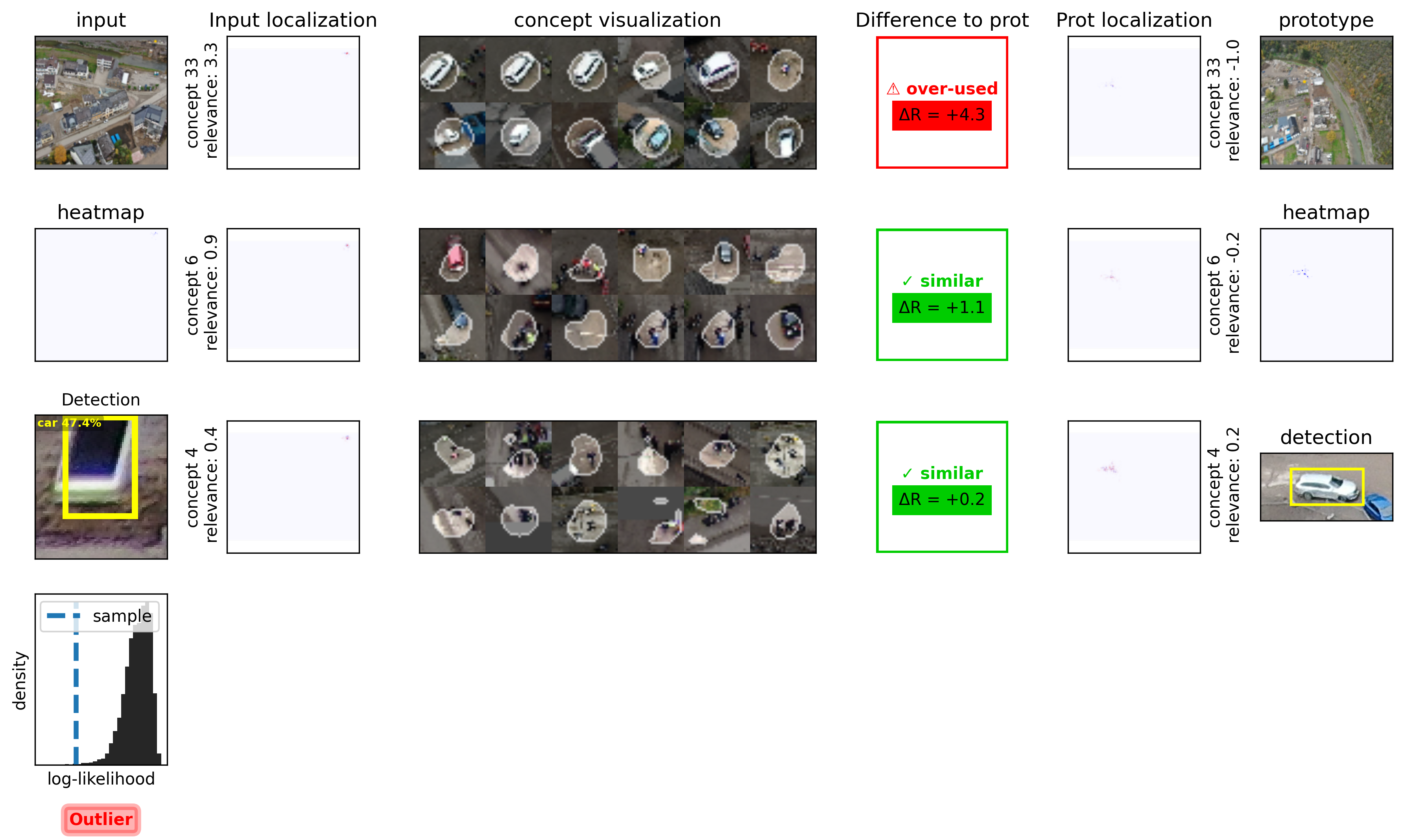}
  \caption{PCX explanation of an outlier prediction for car detection with YOLOv6s6.}\label{fig:car_oulier}
\end{figure}

In Fig.~\ref{fig:car_oulier} we showcase an example prediction from the validation data explained with PCX.
The top left and top right images are the test input and the nearest prototype from the training data. The 3 middle rows correspond to the most relevant concepts used for the prediction, which are visualized by retrieving reference images maximizing each concept (additionally the full heatmap and concept-conditioned heatmaps are provided for each concept). The most important result can be found in the "difference to prototype" column, where concept usage between the test input and the prototype are compared. In particular the top concept, which corresponds to "white vehicle side windows" was over-used in the test input. Hence PCX labeled the prediction as an outlier. Indeed the model mis-detected a window on a roof as a white car, and PCX was able to identify this mistake through the unusual concept profile.

\section{Conclusion}
In this pilot study we demonstrated, both quantitatively and qualitatively, that concept-based explanations can be extended to the NDM domain on two DNN models and tasks in a flood scenario: flood segmentation with PIDNet and car detection with YOLOv6s6. In particular we highlighted that PCX prototypes can help to identify and inspect the model's prediction strategies, and that it can successfully detect outlier predictions. We believe this showcases the usefulness of concept-based XAI for enhancing transparency of DNN decisions in real-world scenarios such as in natural disasters.

\FloatBarrier

\begin{acknowledgments}
  We thank Maximilian Dreyer for helpful discussions. This work was supported by the European Union’s Horizon Europe research and innovation programme's TEMA project, grant number 101093003.
\end{acknowledgments}

\section*{Declaration on Generative AI}
The authors have not employed any Generative AI tools.

\FloatBarrier

\bibliography{bibliography}

@String{Computer = "{IEEE} Computer" }

@INPROCEEDINGS{DreCVPRW23,
  author = {Maximilian Dreyer and Reduan Achtibat and Thomas Wiegand and Wojciech Samek and Sebastian Lapuschkin},
  title = {{Revealing Hidden Context Bias in Segmentation and Object Detection through Concept-specific Explanations}},
  booktitle = {Conference on Computer Vision and Pattern Recognition Workshops},
  year = {2023},
  pages = {3828-3838}
}

@inproceedings{dre2024c,
  title={{Understanding the (Extra-)Ordinary: Validating Deep Model Decisions with Prototypical Concept-based Explanations}},
  author={Maximilian Dreyer and Reduan Achtibat and Wojciech Samek and Sebastian Lapuschkin},
  booktitle={Conference on Computer Vision and Pattern Recognition Workshops},
  year={2024},
  pages = {3491-3501}
}

@article{Fotiou_2025,
title={{RoboFireFuseNet: Robust Fusion of Visible and Infrared Wildfire Imaging for Real-Time Flame and Smoke Segmentation}},
publisher={Institute of Electrical and Electronics Engineers (IEEE)},
journal={TechRxiv},
author={Fotiou, Dimitrios and Mygdalis, Vasileios and Pitas, Ioannis},
year={2025}
}

@article{LI2022145,
title = {A Real-time Fire Segmentation Method Based on A Deep Learning Approach},
journal = {IFAC-PapersOnLine},
volume = {55},
number = {6},
pages = {145-150},
year = {2022},
author = {Mengna Li and Youmin Zhang and Lingxia Mu and Jing Xin and Ziquan Yu and Shangbin Jiao and Han Liu and Guo Xie and Yi Yingmin}
}

@Article{fire8020036,
AUTHOR = {Ali, Abdallah Waleed and Kurnaz, Sefer},
TITLE = {Optimizing Deep Learning Models for Fire Detection, Classification, and Segmentation Using Satellite Images},
JOURNAL = {Fire},
VOLUME = {8},
YEAR = {2025},
NUMBER = {2},
ARTICLE-NUMBER = {36}
}

@Article{fire_scientific_report,
AUTHOR = {Arlovic, Mate and Hrzic, Franko and Patel, Mitesh and Bednarz, Tomasz and Balen, Josip},
TITLE = {Evaluation of synthetic data impact on fire segmentation models performance},
JOURNAL = {Scientific Reports},
VOLUME = {15},
YEAR = {2025},
NUMBER = {16759}
}

@article{Achtibat2023,
author={Achtibat, Reduan
and Dreyer, Maximilian
and Eisenbraun, Ilona
and Bosse, Sebastian
and Wiegand, Thomas
and Samek, Wojciech
and Lapuschkin, Sebastian},
title={{From attribution maps to human-understandable explanations through Concept Relevance Propagation}},
journal={Nature Machine Intelligence},
year={2023},
volume={5},
number={9},
pages={1006-1019}
}

@article{bach2015lrp,
    author = {Bach, Sebastian AND Binder, Alexander AND Montavon, Grégoire AND Klauschen, Frederick AND Müller, Klaus-Robert AND Samek, Wojciech},
    journal = {PLOS ONE},
    publisher = {Public Library of Science},
    title = {On Pixel-Wise Explanations for Non-Linear Classifier Decisions by Layer-Wise Relevance Propagation},
    year = {2015},
    volume = {10},
    pages = {1-46},
    number = {7}
    }

@INPROCEEDINGS{11084431,
  author={Mentesidis, Pantelis and Mygdalis, Vasileios and Pitas, Ioannis},
  booktitle={Int. Conference on Image Processing}, 
  title={Improve Real-Time Flood Segmentation by Encoding and Distilling Foreground Information}, 
  year={2025},
  pages={1840-1845}
  }

@Article{Alqithami2025,
  author = {Alqithami, Saad},
  title = {Integrating Sentiment Analysis and Reinforcement Learning for Equitable Disaster Response: A Novel Approach},
  journal = {Sustainability},
  volume = {17},
  year = {2025},
  number = {3},
  article_number = {1072}
}

@INPROCEEDINGS{11044745,
  author={Gerontopoulos, Anastasios and Papaioannou, Dimitrios and Papaioannidis, Christos and Pitas, Ioannis},
  booktitle={IEEE Int. Symp. on Cluster, Cloud and Internet Comp. Workshops}, 
  title={Real-Time Flood Water Segmentation with Deep Neural Networks}, 
  year={2025},
  pages={85-91}
}

@article{AKHYAR2024112067,
title = {Deep artificial intelligence applications for natural disaster management systems: A methodological review},
journal = {Ecological Indicators},
volume = {163},
pages = {112067},
year = {2024},
author = {Akhyar Akhyar and Mohd {Asyraf Zulkifley} and Jaesung Lee and Taekyung Song and Jaeho Han and Chanhee Cho and Seunghyun Hyun and Youngdoo Son and Byung-Woo Hong}
}

@Article{make4020020,
AUTHOR = {Linardos, Vasileios and Drakaki, Maria and Tzionas, Panagiotis and Karnavas, Yannis L.},
TITLE = {Machine Learning in Disaster Management: Recent Developments in Methods and Applications},
JOURNAL = {Mach. L. and Knowl. Extract.},
VOLUME = {4},
YEAR = {2022},
NUMBER = {2},
PAGES = {446--473}
}

@INPROCEEDINGS{PIDNet,
  author={Xu, Jiacong and Xiong, Zixiang and Bhattacharyya, Shankar P.},
  booktitle={CVPR}, 
  title={{PIDNet: A Real-time Semantic Segmentation Network Inspired by PID Controllers}}, 
  year={2023},
  volume={},
  number={},
  pages={19529-19539}
}

@INPROCEEDINGS{11084700,
  author={Siavrakas, Michael and Papaioannidis, Christos and Pitas, Ioannis},
  booktitle={Int. Conference on Image Processing}, 
  title={{Blaze: A Dataset For Wildfire And Burnt Area UAV Image Classification And Segmentation}}, 
  year={2025},
  pages={1960-1965}
}

@article{PI2020101009,
title = {Convolutional neural networks for object detection in aerial imagery for disaster response and recovery},
journal = {Advanced Engineering Informatics},
volume = {43},
pages = {101009},
year = {2020},
author = {Yalong Pi and Nipun D. Nath and Amir H. Behzadan}
}

@Article{smartcities8010012,
AUTHOR = {Zhao, Zi-An and Wang, Shidan and Chen, Min-Xin and Mao, Ye-Jiao and Chan, Andy Chi-Ho and Lai, Derek Ka-Hei and Wong, Duo Wai-Chi and Cheung, James Chung-Wai},
TITLE = {{Enhancing Human Detection in Occlusion-Heavy Disaster Scenarios: A Visibility-Enhanced DINO (VE-DINO) Model with Reassembled Occlusion Dataset}},
JOURNAL = {Smart Cities},
VOLUME = {8},
YEAR = {2025},
NUMBER = {1},
ARTICLE-NUMBER = {12}
}

@article{owid-natural-disasters,
    author = {Hannah Ritchie and Pablo Rosado and Max Roser},
    title = {Natural Disasters},
    journal = {Our World in Data},
    year = {2022},
    url = {https://ourworldindata.org/natural-disasters}
}

@article{arras2017lstm,
  author  = {Arras, Leila and Montavon, Gr{\'e}goire and Klauschen, Frederick and M{\"u}ller, Klaus-Robert and Samek, Wojciech},
  title   = {Explaining Recurrent Neural Network Predictions in Sentiment Analysis},
  journal = {EMNLP Workshop on Computational Approaches to Subjectivity, Sentiment and Social Media Analysis},
  year    = {2017}
}

@article{arras2025,
  author  = {Arras, Leila and Puri, Bruno and Kahardipraja, Patrick and Lapuschkin, Sebastian and Samek, Wojciech},
  title   = {A Close Look at Decomposition-based {XAI}-Methods for {Transformer} Language Models},
  journal = {arXiv:2502.15886},
  year    = {2025}
}

@article{Li2024YOLOv6,
  author  = {Chuyi Li and Lulu Li and Hongliang Jiang and Kaiheng Weng and Yifei Geng and et al.},
  title   = {{YOLOv6: A Single-Stage Object Detection Framework for Industrial Applications}},
  journal = {arXiv.2209.02976},
  year    = {2022}
}

@inproceedings{spatharis2025unrealfire,
  title={{Unrealfire: A Synthetic Dataset Creation Pipeline for Annotated Fire Imagery in Unreal Engine}},
  author={Spatharis, Evangelos and Papaioannidis, Christos and Mygdalis, Vasileios and Pitas, Ioannis},
  booktitle={Int. Conference on Image Processing Workshops},
  pages={610--615},
  year={2025}
}

@article{tzimas2026extreme,
  title={Extreme weakly supervised binary semantic image segmentation via one-pixel supervision},
  author={Tzimas, Matthaios and Mygdalis, Vasileios and Papaioannidis, Christos and Pitas, Ioannis},
  journal={Pattern Recognition},
  pages={113048},
  year={2026}
}

@article{APOSTOLIDIS2026133290,
title = {{MEWS: Semantic image segmentation with multiclass extreme weak supervision}},
author = {A. Apostolidis and V. Mygdalis and M. Tzimas and I. Pitas},
journal = {Neurocomputing},
pages = {133290},
year = {2026}
}

@INPROCEEDINGS{11226416,
  author={Vlachos, Evgenios and Papaioannidis, Christos and Pitas, Ioannis},
  booktitle={EUSIPCO}, 
  title={Neural Architecture Search and Knowledge Distillation for Semantic Image Segmentation on Big Wildfire Datasets}, 
  year={2025},
  volume={},
  number={},
  pages={750-754}
}

@article{wieland2025fusion,
  title={Fusion of geospatial information from remote sensing and social media to prioritise rapid response actions in case of floods},
  author={Wieland, Marc and Schmidt, Sebastian and Resch, Bernd and Abecker, Andreas and Martinis, Sandro},
  journal={Natural Hazards},
  volume={121},
  number={7},
  pages={8061-8088},
  year={2025}
}

@InProceedings{Selvaraju:ICCV2017,
author = {Selvaraju, Ramprasaath R. and Cogswell, Michael and Das, Abhishek and Vedantam, Ramakrishna and Parikh, Devi and Batra, Dhruv},
title = {{Grad-CAM}: Visual Explanations from Deep Networks via Gradient-Based Localization},
booktitle = {ICCV},
year = {2017}
}

@inproceedings{Simonyan:ICLR2014,
  author    = {Karen Simonyan and Andrea Vedaldi and Andrew Zisserman},
  title     = {Deep Inside Convolutional Networks: Visualising Image Classification Models and aliency Maps},
  booktitle = {ICML},
  year      = {2014}
}

@InProceedings{Lundberg:NIPS2017,
title = {A Unified Approach to Interpreting Model Predictions},
author = {Lundberg, Scott M. and Lee, Su-In},
booktitle = {Advances in Neural Information Processing Systems},
pages = {4765--4774},
year = {2017}
}

@article{arras2022clevr,
  title={{CLEVR-XAI: A benchmark dataset for the ground truth evaluation of neural network explanations}},
  author={Arras, Leila and Osman, Ahmed and Samek, Wojciech},
  journal={Information Fusion},
  volume={81},
  pages={14--40},
  year={2022}
}

@article {Mamalakis2022,
      author = "Antonios Mamalakis and Elizabeth A. Barnes and Imme Ebert-Uphoff",
      title = {Investigating the Fidelity of Explainable Artificial Intelligence Methods for Applications of Convolutional Neural Networks in Geoscience},
      journal = "Artificial Intelligence for the Earth Systems",
      year = "2022",
      volume = "1",
      number = "4"
}

@inproceedings{KimWGCWVS18,
  author = {Kim, Been and Wattenberg, Martin and Gilmer, Justin and Cai, Carrie J. and Wexler, James and Viégas, Fernanda B. and Sayres, Rory},
  booktitle = {ICML},
  title = {Interpretability Beyond Feature Attribution: Quantitative Testing with Concept Activation Vectors {(TCAV)}},
  year = 2018
}

@incollection{Montavon:ExplAIBook2020,
  author    = {Gr{\'{e}}goire Montavon and Alexander Binder and Sebastian Lapuschkin and Wojciech Samek and Klaus-Robert M{\"u}ller},
  title     = {Layer-Wise Relevance Propagation: An Overview},
  booktitle = {Explainable {AI}: Interpreting, Explaining and Visualizing Deep Learning},
  volume = {11700},
  pages = {193-209},
  year      = {2019},
  series    = {LNCS}
}

@ARTICLE{samek_2017,
  author={Samek, Wojciech and Binder, Alexander and Montavon, Grégoire and Lapuschkin, Sebastian and Müller, Klaus-Robert},
  journal={IEEE TNNLS}, 
  title={Evaluating the Visualization of What a Deep Neural Network Has Learned}, 
  year={2017},
  volume={28},
  number={11},
  pages={2660-2673}
  }

@INPROCEEDINGS{Kohlbrenner_2020,
  author={Kohlbrenner, Maximilian and Bauer, Alexander and Nakajima, Shinichi and Binder, Alexander and Samek, Wojciech and Lapuschkin, Sebastian},
  booktitle={Int. Joint Conf. on Neural Netw.}, 
  title={Towards Best Practice in Explaining Neural Network Decisions with {LRP}}, 
  year={2020},
  volume={},
  number={},
  pages={1-7}
  }

\end{document}